\documentclass[conference]{IEEEtran}
\IEEEoverridecommandlockouts
\usepackage{cite}
\usepackage{amsmath,amssymb,amsfonts}
\usepackage{algorithmic}
\usepackage{graphicx}
\usepackage{textcomp}
\usepackage{xcolor}
\usepackage{subfigure}
\def\BibTeX{{\rm B\kern-.05em{\sc i\kern-.025em b}\kern-.08em
    T\kern-.1667em\lower.7ex\hbox{E}\kern-.125emX}}
\begin{document}

\title{Reward-Punishment Reinforcement Learning with Maximum Entropy\\

}

\author{\IEEEauthorblockN{Jiexin Wang}
\IEEEauthorblockA{\textit{Dept. Brain Robot Interface} \\
\textit{ATR Computational Neuroscience Laboratories}\\
Kyoto, Japan \\
wang-j@atr.jp}
\and
\IEEEauthorblockN{Eiji Uchibe}
\IEEEauthorblockA{\textit{Dept. Brain Robot Interface} \\
\textit{ATR Computational Neuroscience Laboratories}\\
Kyoto, Japan \\
uchibe@atr.jp}
\\}

\maketitle

\begin{abstract}
We introduce the ``soft Deep MaxPain'' (softDMP) algorithm, which integrates the optimization of long-term policy entropy into reward-punishment reinforcement learning objectives. Our motivation is to facilitate a smoother variation of operators utilized in the updating of action values beyond traditional ``max'' and ``min'' operators, where the goal is enhancing sample efficiency and robustness. We also address two unresolved issues from the previous Deep MaxPain method. Firstly, we investigate how the negated (``flipped'') pain-seeking sub-policy, derived from the punishment action value, collaborates with the ``min'' operator to effectively learn the punishment module and how softDMP's smooth learning operator provides insights into the ``flipping'' trick. Secondly, we tackle the challenge of data collection for learning the punishment module to mitigate inconsistencies arising from the involvement of the ``flipped'' sub-policy (pain-avoidance sub-policy) in the unified behavior policy. We empirically explore the first issue in two discrete Markov Decision Process (MDP) environments, elucidating the crucial advancements of the DMP approach and the necessity for soft treatments on the hard operators. For the second issue, we propose a probabilistic classifier based on the ratio of the pain-seeking sub-policy to the sum of the pain-seeking and goal-reaching sub-policies. This classifier assigns roll-outs to separate replay buffers for updating reward and punishment action-value functions respectively. Our framework demonstrates superior performance in Turtlebot 3's maze navigation tasks under the ROS Gazebo simulation. 

\end{abstract}


\section{Introduction}
\label{sec:introduction}
Rewards and punishments specify the goal of learning agents in Reinforcement Learning (RL), and reward-punishment RL constitutes a dichotomic framework that divides standard RL with its monolithic reward structure into positive and negative counterparts. 
Elfwing and Seymour proposed the MaxPain architecture \cite{Elfwing2017a}, inspired by the separation of reward and pain mechanisms observed in the brain activity of animals \cite{Seymour2007a,Seymour2012a,Eldar2016a}. 
A key feature of MaxPain is its dual objectives: maximizing the expectations of positive returns while simultaneously minimizing the expectations of negative returns. Namely, MaxPain maintains two action-value functions: $Q_+$ for positive rewards and $Q_-$ for negative rewards or punishments, from which a reward-seeking sub-policy and a punishment-seeking one can be derived respectively. A composite value function combining $Q_+$ and $-Q_-$ was obtained for the action selection where the latter is the negated value function of $Q_-$.
Deep MaxPain (DMP) extended MaxPain by incorporating deep learning techniques \cite{Wang2017b, Wang2021a}. Unlike MaxPain, DMP constructs a composite policy from the reward-seeking sub-policy $\pi_+ (a | s) \propto \exp (\eta_+ Q_+ (s, a))$ and the pain-avoiding sub-policy $\neg \pi_- (a | s) \propto \exp (- \eta_- Q_- (s, a))$, where $\eta_+$ and $\eta_-$ serve as hyperparameters. Here, the pain-avoiding sub-policy $\neg \pi_-$ is referred to as the ``flipped'' policy, in correspondence with the pain-seeking sub-policy $\pi_- (a | s) \propto \exp (\eta_- Q_- (s, a))$.


While DMP showed its effectiveness empirically, two open questions remain. 
First, it is unclear whether the flipped policy $\neg \pi_-$ is superior to the standard policy $\pi_-$ for avoiding negative rewards. To address this issue, we delve into the critical nature of the ``flipped'' policy, and show how it collaborates with the ``min'' operator for learning a punishment module to achieve superior performance compared to using the ``max'' operator. We systematically explore smoothed operators for action-value optimization by leveraging an entropy-regularized RL \cite{Asadi2017a}. 
Combining a max-entropy regularizer with a reward-punishment framework gives us a soft Deep MaxPain (soft DMP) algorithm, which showcases the interplay between the entropy coefficient and various operators, encompassing ``max,'' ``mellow-max,'' ``mean,'' ``mellow-min'' and ``min'', and provides appealing features for improving sample efficiency and robustness. 

The second question: if the ``flipping'' is affable, how should the data for training $Q_-$ should be properly collected? Even though learning $Q_-$ is an off-policy process, the experience gathered by the composite policy might be inappropriate since $\neg \pi_-$ differs from the optimal policy $\pi_-^*$ derived from $Q_-^*$. 
That is, training $Q_-$ with experience collected by a composite policy is sample inefficient because the buffer is occupied by reward-seeking and pain-avoiding data. This situation resembles the setting of offline reinforcement learning. To tackle this, we prepared two experience replay buffers and classified the experiences by an optimal discriminator, inspired by Generative Adversarial Networks (GANs) \cite{Goodfellow2014a}. Our discriminator is given by a probability directly obtained from $\pi_+$ and $\pi_-$, eliminating the need for explicit training. The discriminator then assigns experience tuples to two separate replay buffers. An experience labeled as ``negative'' is deliberately allocated to update the punishment action value. 


We investigated the impact of the entropy coefficients in non-positive reward settings by conducting two simple discrete Markov Decision Problems: Grid-world and Chain environments. 
Numerical simulations revealed that negative rewards are not effectively propagated when the coefficient is non-negative. On the other hand, negative rewards are propagated over the state space when the coefficient is negative, indicating that minimizing the sum of negative rewards is more informative than maximizing it. We further evaluated softDMP with our proposed discriminator for separating a replay buffer on vision-based navigation tasks with Turtlebot3 in an ROS Gazebo environment. 
Experimental results demonstrated improved performance compared to utilizing only one buffer.


Our key contributions include:
\begin{itemize}
\item a maximum-entropy based Deep MaxPain algorithm (softDMP), from which the entropy parameter governs a spectrum of operators providing the flexibility for fine adjustment to utilize the learned Q values

\item an empirical analysis on the significance of ``min'' or ``mellow-min'' operators with a ``flipped'' policy for learning a negative reward environment, which essentially enhances the DMP performance 

\item a separate replay buffer scheme on softDMP, to address the inconsistency problem arising from the ``flipping'' trick in learning the negative reward, and improving data efficiency and robustness in maze-solving navigation tasks.
\end{itemize}

These contributions collectively enhance our understanding and application of softDMP algorithms, showcasing their adaptability and effectiveness in complex learning scenarios.
 
\section{Related Work}

\subsection{Separating Reward and Punishment}

The reward-punishment framework fundamentally represents a dual-structured value decomposition practice \cite{cql,gmq,hra}, situated within Modular Reinforcement Learning (MoRL) \cite{humphrys1996,doya2002}. Its motivation is to break a complex monolithic task into several simpler sub-tasks, each of which is learned in parallel; the original problem is tackled after recomposing the resulting modules. The dichotomous decomposition is traced back to a two-dimensional evaluation RL \cite{okada2001two}. In this framework, the TD-error learned from each sub-module is utilized for both exploring and correcting the preference value in the Actor-Critic method. A similar approach is found in a meta-learning model \cite{lowe2013exploring}, where each TD-error is used to update a hyper-action structure in dyna-SARSA. 

From a biological perspective, recent neuroscience studies elucidate that separate reward and punishment systems function in animal brain activities \cite{Seymour2007a,Seymour2012a,Eldar2016a}. MaxPain \cite{Elfwing2017a} investigated the feasibility of such a dualistic structure in classic Q-learning and SARSA algorithms, emphasizing advances in terms of exploration and safety. Maxpain's learning mechanism is akin to that of the Hybrid Reward Architecture \cite{hra}, where the distinction is that the punishment (negative-reward) sub-module is learned by a ``min'' Bellman operator. DMP \cite{Wang2017b,Wang2021a} argued for the necessity of such a learning rule in back-propagating the true amount of negative signals, while explaining how pain-avoidance induces sample efficiency and exploration. RP-AC \cite{rlac} is a revamped reward-punishment Actor-Critic framework that formulates a policy gradient for continuous control. Split-Q Learning \cite{split} suggested a more generalized reward-punishment framework by parameterizing immediate positive and negative rewards and their approximated state-action-values, aligning with various neurological and psychiatric mechanisms.

Other than SARSA and Actor-Critic, DMPs \cite{Wang2017b}\cite{Wang2021a} construct an off-policy discrete action framework based on MaxPain. Both the reward and punishment modules are learned in a Q-learning manner, except the punishment module is learned with the ``min'' operator. However, DMP does not elucidate the underlying principle behind the usage of the ``flipped'' pain-seeking sub-policy, which is actually the pain-avoidance sub-policy, and why its collaboration with the ``min'' operator in learning the punishment module improves learning performance. In this study, we first empirically clarify this issue in a low-dimensional Grid-world task. We then augment DMPs with the maximum-entropy constraint and demonstrate how the entropy parameter can alleviate the greediness of the ``max'' and ``min'' operators in learning negative signals. In a Chain environment, we see that different magnitudes of the negative rewards can be back-propagated with the smoothed operators and the participation of the ``flipped'' pain-seeking sub-policy. softDMP also achieved superior learning performance in Turtlebot's navigation tasks.

\subsection{Maximum Entropy Reinforcement Learning}

The maximum-entropy principle, which has been incorporated into the deep reinforcement learning literature for a number of successful achievements, introduces an entropy of policy in the objective functions to encourage future stochasticity of the learning policy. Namely, the policy entropy is maximized alongside the expected return. This approach enables the agent to achieve better exploration and avoid being trapped in local optima, ensuring improved convergence in terms of data efficiency.

Moreover, in the context of discrete actions, maximizing the entropy bonus can be interpreted as minimizing a Kullback-Leibler (KL) regularizer between the learning policy and a prior policy \cite{Fox2016a,Azar2012a,Toussaint2009a}. 
KL divergence serves as a statistical metric to measure the proximity of two probability distributions. 
As a penalty term, this metric motivates the learning policy to closely align with the prior in the learning process. G-learning \cite{Fox2016a}, which is one classical method in KL-regularized RL, emphasized that the baseline policy in Dynamic Policy Programming \cite{Azar2012a} can be a uniform prior in the model-free off-policy RL domain, particularly coping with environment noise in the early stage of learning. 
The updating rule under such regularization therefore can be derived into a log-average-exp operator known as ``mellow-max'' \cite{Asadi2017a}. 
This operator boasts appealing properties, including non-expansion to guarantee fixed-point convergence and differentiability with respect to the entropy parameter \cite{Asadi2017a}. The entropy parameter, also known as the parameter for the KL regularizer, calibrates the extent to which the entropy term participates in the objective. It plays a crucial role in mitigating the greediness of the traditional hard-max operator for updating the action values.

Soft Q-Learning (SQL) \cite{Haarnoja2017a} and Soft Actor-Critic \cite{Haarnoja2018a} inherited the KL constraint. Here the maximum entropy constraint on the learning policy, similar to the methods mentioned above, achieved state-of-the-art results in deep reinforcement learning for continuous control. See a previous work \cite{Levine2018a} for a summary that relates the above two methods to the probabilistic graphical models. In addition, two other works \cite{Kozuno2019a} and \cite{Eysenbach2022a} provided theoretical analyses of how soft (maximum-entropy RL) approaches are tolerant to noise and biases, thereby inducing convincing robustness properties. 

Our method mainly takes a G-learning formulation \cite{Fox2016a}, and extends it into a reward-punishment setup. In G-learning, entropy parameter $\eta$ should be positive, which is represented as ``mellow-max'' for learning a complete reward structure. In our method, $\eta$ covers the real number domain: from positive infinity to negative infinity, corresponding to the range from ``max'' to ``min'' operators for action-value optimizations. The flexibility of $\eta$ can be applied, for example, either in learning a solely negative reward with ``max'' to ``min'' operators or in learning two separate modules like positive $\eta$ for the reward module and negative $\eta$ for the punishment module. The smoothness of $\eta$ provides a systematic way not only to evaluate the effectiveness of a learning punishment module, but also to improve the learning of a positive reward module, which essentially enhances the performance of softDMPs.

\section{Preliminaries: Soft Q-learning}
\label{sec:SQL}

We consider an infinite discrete-time Markov Decision Process (MDP) defined by a tuple $(\mathcal{S}, \mathcal{A}, \mathcal{P}, \mathcal{P}_0, \mathcal{R}, \gamma)$, where $\mathcal{S}$ is the state space, $\mathcal{A}$ is the discrete action set, $\mathcal{P}$ is the state transition probability, $\mathcal{P}_0$ is the initial state distribution, $\mathcal{R}$ is the reward function, and $\gamma$ is the discount factor. 
First, an initial state is determined by $s_0 \sim \mathcal{P}_0$.
At each time step $t$, an agent takes action $a_t \in \mathcal{A}$ in response to current state $s_t \in \mathcal{S}$, depending on a stochastic policy $\pi (a_t|s_t)$. 
Subsequently, the environment transits the agent to subsequent state $s_{t+1} \sim \mathcal{P}(\cdot | s_t, a_t)$, yielding a scalar reward from deliberate reward function $r_t = \mathcal{R}(s_t,a_t,s_{t+1})$ as an evaluation signal. 

In Soft Q-learning, the goal is to maximize the expected discounted return along with an entropy regularization term, which is given by a KL regularizer between the learning policy and a uniform random policy:
\begin{equation}
\label{eq:obj}
\max_{\pi} \mathbb{E}_{s \sim \mathcal{P}_0} [\mathrm{sign}(\eta) V_\eta^\pi (s)],
\end{equation}
where $V^\pi(s)$ is a soft state-value function defined by
\begin{align}
& V_\eta^\pi(s) = \notag \\
& \mathbb{E}_{\pi, \mathcal{P}} \left[ \sum_{t=0}^\infty \gamma^t \left( r_t - \frac{1}{\eta} KL (\pi (\cdot | s_t) \parallel \pi_u (\cdot | s_t) \right) | s_0 = s \right], \notag
\end{align}
where $\eta \in (-\infty, \infty)$ is an entropy parameter, $\pi_u(a) = 1 / |\mathcal{A}|$ is a uniform policy, $\gamma \in [0,1)$ is a discount factor that determines how far in the future to look for rewards, and $\pi_u$ denotes a uniform random prior. 
The KL term can be interpreted as a penalty if the learning policy deviates from the uniform prior, that is, to encourage a more stochastic policy than standard Q-learning. 
Note that we multiply soft state-value function by $\eta$ in Eq.~(\ref{eq:V}) for the later purpose.

Applying the Lagrangian multiplier method yields an optimal state value function \cite{Fox2016a,Asadi2017a,Haarnoja2017a}:
\begin{equation}
\label{eq:V}
  V_\eta^*(s) =\mathrm{MM}_\eta Q_\eta^*(s, a),
\end{equation}
where $Q_\eta^*(s, a)$ is an optimal soft action-value function, and $\mathrm{MM}_\eta$ is a mellow-max operator \cite{Asadi2017a} defined by
\begin{align}
  \label{eq:MM_operator}
  \mathrm{MM}_{\eta} Q (s, a) \triangleq \frac{1}{\eta} \log \sum_{a \in \mathcal{A}} \frac{1}{|\mathcal{A}|} \exp \left[\eta Q (s, a) \right].
\end{align}
The right-hand side of Eq.~(\ref{eq:MM_operator}) is a log-average-exp function and generalizes an operator as follows:
\begin{equation}
\label{eq:mmq}
\mathrm{MM}_\eta Q(s, a) =
\begin{cases}
  \max_a Q(s, a) & \eta = + \infty \\
  \mathrm{mellow} \max_a Q(s, a) & \eta \in (0, + \infty) \\
  \mathrm{mean}_a Q(s, a) & \eta = 0 \\
  \mathrm{mellow} \min_a Q(s, a) & \eta \in (-\infty, 0) \\
  \min_a Q(s, a) & \eta = - \infty.
\end{cases}
\end{equation}

When the action-value function is represented by a look-up table, $Q^*$ is obtained by applying the following learning rule for the experience $(s_t, a_t, r_t, s_{t+1})$:
\begin{displaymath}
  Q_\eta (s_t, a_t) \leftarrow (1 - \alpha)Q_\eta (s_t, a_t) + \alpha (r_t + \gamma \mathrm{MM}_{\eta} Q_\eta (s_{t+1}, \cdot) ),  
\end{displaymath}
where $\alpha$ is a positive learning rate, 

The property of $\mathrm{MM}_\eta$ for $\eta \in [0, +\infty)$ was previously reported \cite{Asadi2017a}, although we expand $\eta$ because it is useful to systematically treat the MaxPain architecture.

\section{Proposed Method: softDMP}
\label{sec:softDMP}

\subsection{Generalization of DMP's Objective Functions}

Based on Soft Q-Learning for $\eta \in (-\infty, +\infty)$, we can rewrite MaxPain's objective functions.
For the negative reward,
\begin{displaymath}
  r^- = \min(\mathcal{R},0) \leq 0,
\end{displaymath}
the DMP minimizes
\begin{displaymath}
  \min_{\pi_-} \mathbb{E}_{s \sim \mathcal{P}_0}
  \left[ V_{-\infty}^\pi (s) \right]
  = \max_{\pi_-} \mathbb{E}_{s \sim \mathcal{P}_0} \left[ \mathrm{sign}(-\infty) V_{-\infty} (s) \right].
\end{displaymath}
Similarly, for the positive reward
\begin{displaymath}
  r^+ = \max(\mathcal{R}, 0) \geq 0,
\end{displaymath}
DMP maximizes
\begin{displaymath}
  \max_{\pi_+} \mathbb{E}_{s \sim \mathcal{P}_0}
  \left[ V_{+\infty}^\pi (s) \right]
  = \max_{\pi_+} \mathbb{E}_{s \sim \mathcal{P}_0} \left[ \mathrm{sign}(+\infty) V_{+\infty} (s) \right].
\end{displaymath}

Based on the above observation, softDMP updates the value function for the positive and negative rewards.
Note that $\eta_+ \in (0, +\infty)$ is paired with $Q_+(>0)$ for ``maximizing'' and ``mellow-maximizing'' the positive reward module, while $\eta_- \in (-\infty, 0)$ is paired with $Q_-(<0)$ for ``minimizing'' and ``mellow-minimizing'' the negative one. When $\eta = 0$, it simply stands for the ``mean'' operator in action-value optimization.

\subsection{Behavior Policy Fusion}

We define the sub-policies derived from two action values:
\begin{align}
  \pi_+(a|s) &\propto \exp [\eta_+Q_+(s,a)], 
  \quad \eta_+\geq 0, \notag \\
  \pi_-(a|s) &\propto \exp [\eta_-Q_-(s,a)], 
  \quad \eta_-\leq 0, \notag
\end{align}
where $\pi_+(a|s)$ is the target-approaching sub-policy from learning the positive reward module, and $\pi_-(a|s)$ is the pain-approaching sub-policy from the negative one. We also define a ``flipped'' negative sub-policy, which serves as a pain-avoiding sub-policy:
\begin{displaymath}
\neg \pi_-(a|s) \propto \exp [-\eta_-Q_-(s,a)], 
\quad \eta_- \leq 0.
\end{displaymath}
The negated negative-policy is then combined with its companion positive-policy to form a unified behavior policy
\begin{displaymath}
\bar{\pi} (a | s) = w \pi_+(a|s) + (1-w) \neg \pi_-(a|s),
\end{displaymath}
where $w \in [0, 1]$ is a mixing weight. 
This enables the target-approaching sub-policy to collaborate with the pain-avoidance sub-policy for more efficient parallel learning. 

\subsection{Separate Replay Buffer}

In the model-free learning case, a challenge arises when negated sub-policy $\neg \pi_-(a|s)$ is included in the behavior policy. The experience gathered by $\neg \pi_-(a|s)$ is meant to avoid pain, which contradicts the objective of updating negative action-value $Q_-(s,a)$ aimed at learning the most painful signal. This contradiction creates a gap preventing $Q_-(s,a)$ from converging to its optimal value. Such an issue has been examined in the context of offline RL literature. In our approach, we address it by introducing a separate replay buffer.

To bridge the gap between the behavior policy and the update of negative action values, we introduce a policy-dependent probabilistic classifier whose formulation is inspired by the discriminator in Generative Adversarial Networks \cite{Goodfellow2014a}.
Suppose that $p_+ (a, s', r | s)$ and $p_-(a, s', r | s)$ denote the probability distributions connected to $\pi_+$ and $\pi_-$. 
For example, $p_+(a, s', r | s) = \mathcal{P}(s' | s, a) \pi_+ (a | s)$ because $r$ is assumed to be a deterministic function.
The loss function of the discriminator is given by the following negative log-likelihood:
\begin{equation}
\label{eq:loss_discriminator}
J(D) = - \mathbb{E}_{p_-} [\ln D(s, a, r, s')] -
\mathbb{E}_{p_+} [\ln (1 - D(s, a, r, s')) ].
\end{equation}
Then the optimal discriminator that minimizes $J(D)$ is given by
\begin{displaymath}
D(s, a, r, s') = \frac{\pi_-(a|s)}{\pi_-(a|s)+\pi_+(a|s)},
\end{displaymath}
where $D(s, a, r, s')$ represents the probability that the experience tuple $(s, a, r, s')$ is generated by $\pi_-$. 
The optimal discriminator assigns the experience tuple with probability $D(s, a, r, s')$ to the negative replay buffer $\mathcal{D}_-$ to update $Q_-$, and the remaining tuples enter positive replay buffer $\mathcal{D}_+$ to update $Q_+$.

\section{Experiments}

\subsection{Grid-world}

We begin by demonstrating how the ``min'' operator in conjunction with a ``flipped'' policy--a policy derived from its negated action value--facilitates a more meaningful learning of pain-avoidance, compared to the usage of the ``max'' operator with a ``non-flipped'' policy. We employed Q Value Iteration (QVI): a model-based approach in a 9x9 U-maze Grid-world. The goal state is on the right side next to an inner wall, and the starting state is on the left side next to the inner wall. The experimental setup aligns with that described in Section 4.1.1 \cite{Wang2021a}, except that the agent only gains negative signals $r=-0.1$ when colliding with a wall. Specifically, the agent receives 0 reward everywhere else, including the absorbing-goal state. We added a ``stop'' action to the four-directional action space.

\begin{figure}
    \centering
    \subfigure[``max'' operator]
    {
        \includegraphics[width=0.46\hsize]{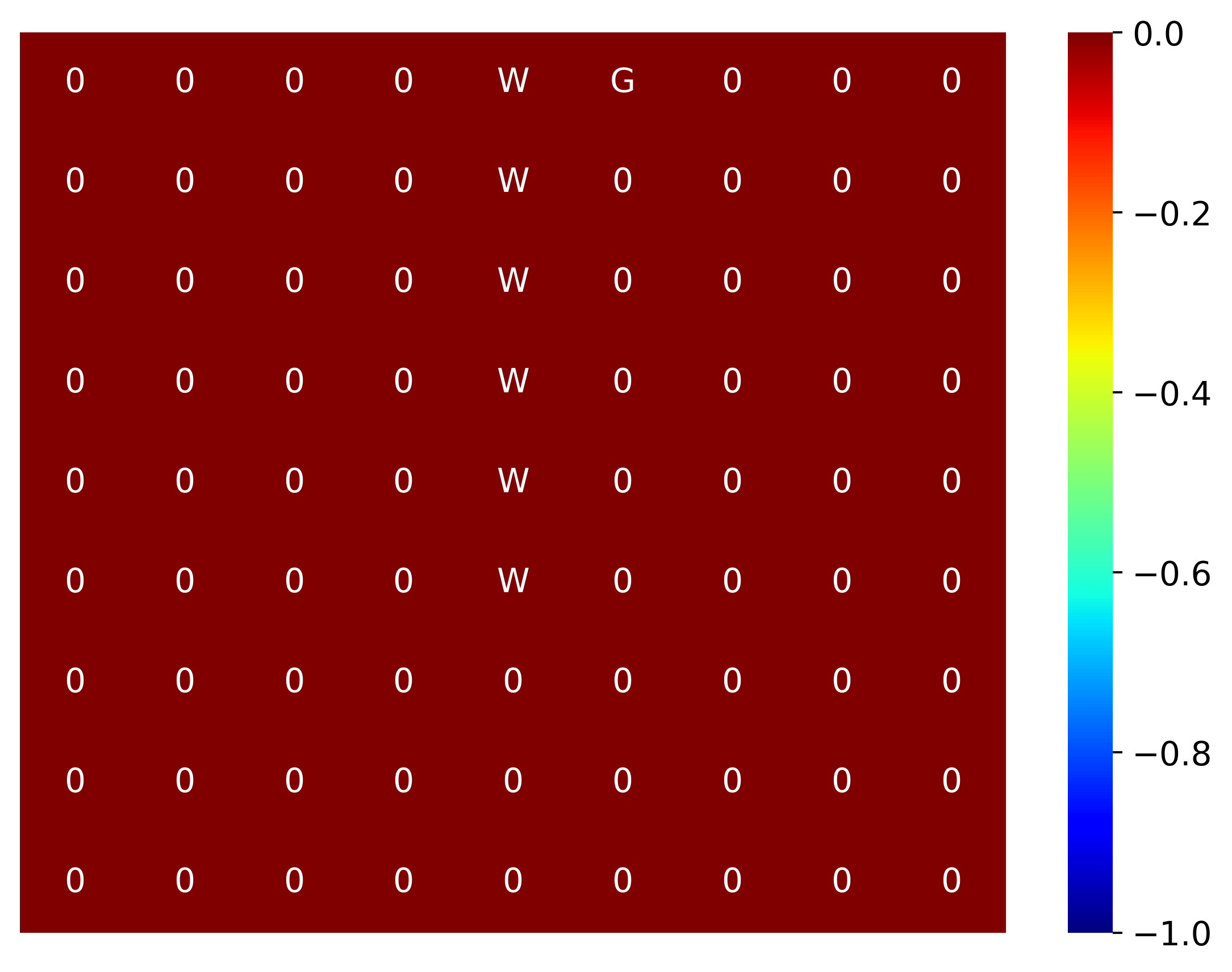}
        \label{fig:max}
    }
    \subfigure[``min'' operator]
    {
        \includegraphics[width=0.46\hsize]{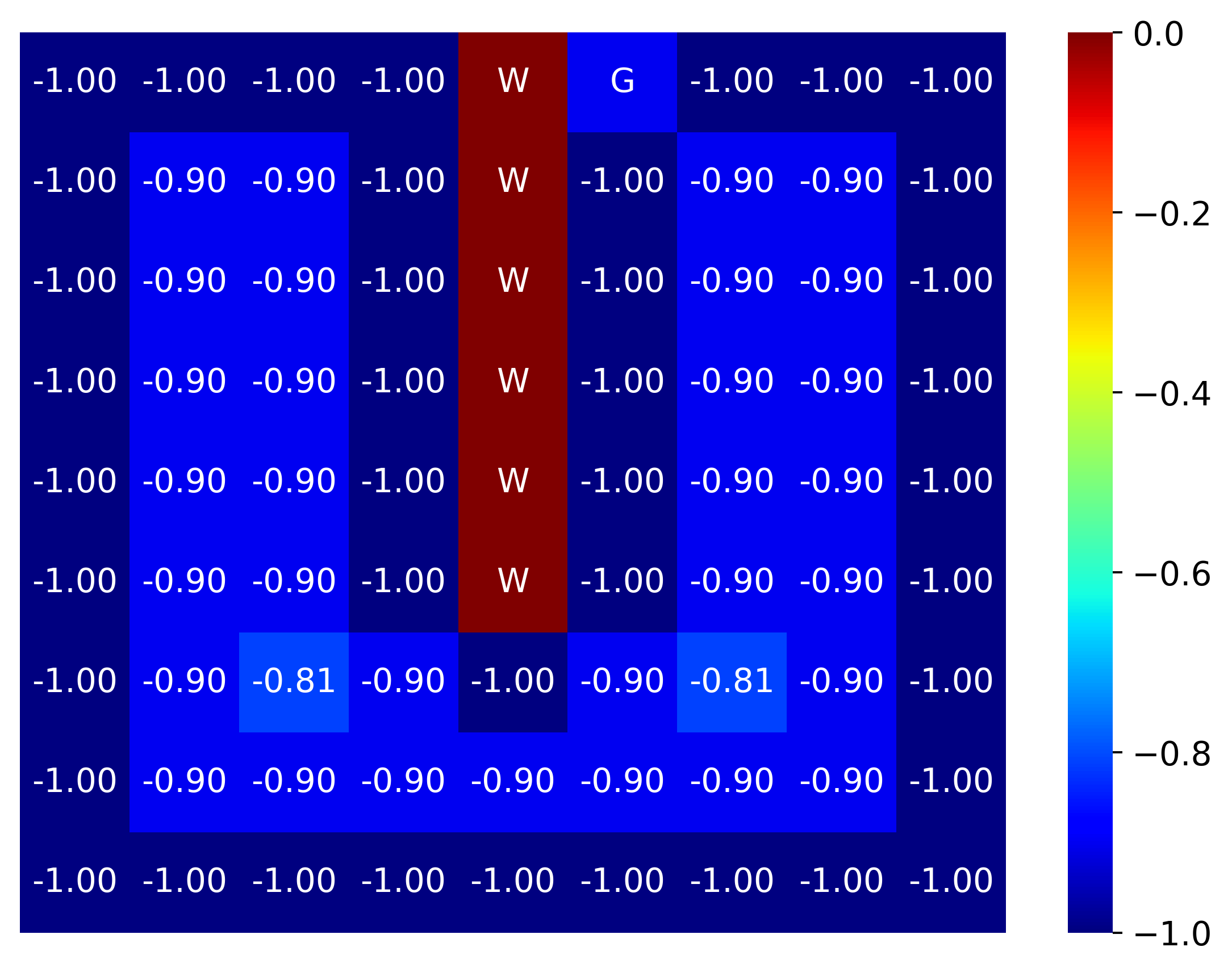}
        \label{fig:min}
    }
    
    \caption{Optimal state value $V^*$ learned by ``max'' and ``min'' operators with Q Value Iteration in 9x9 U-maze Grid-world (Note that black bar signifies ``stop'' action)}
    \label{fig1}
    
\end{figure}

\begin{figure}
    \centering
    \subfigure[$\arg\max(Q^{max})$]
    {
        \includegraphics[width=0.46\hsize]{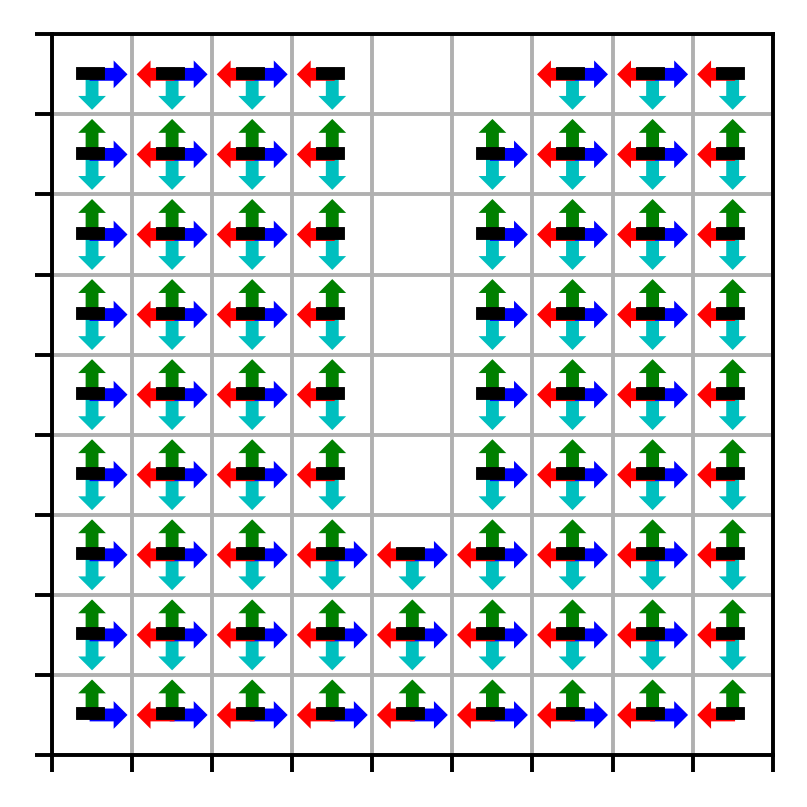}
        \label{fig:2a}
    }
    \subfigure[$\arg\min(Q^{max})$]
    {
        \includegraphics[width=0.46\hsize]{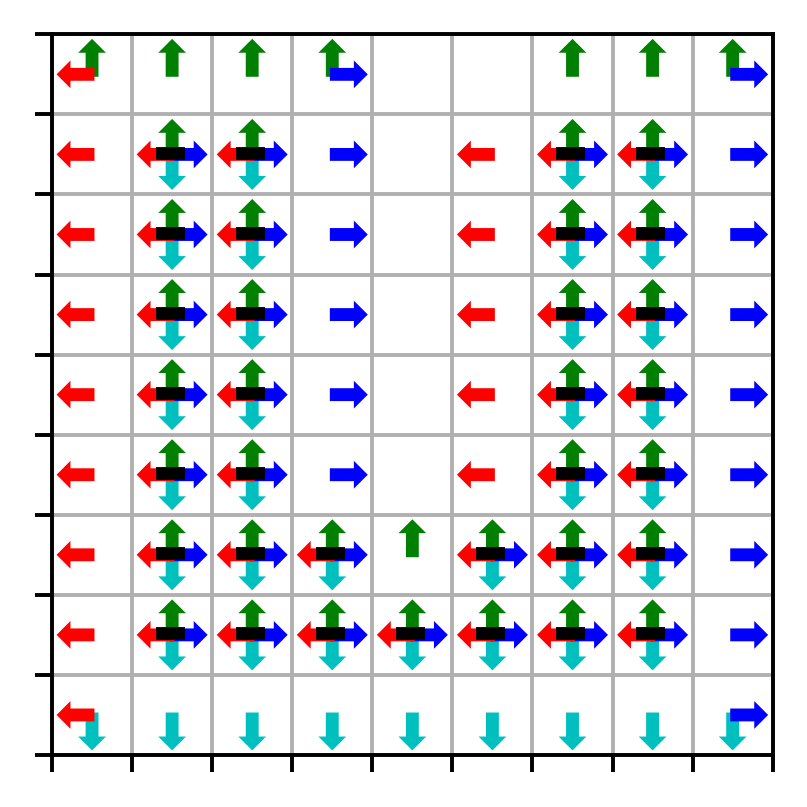}
        \label{fig:2b}
    }

    \subfigure[$\arg\max(Q^{min})$]
    {
        \includegraphics[width=0.46\hsize]{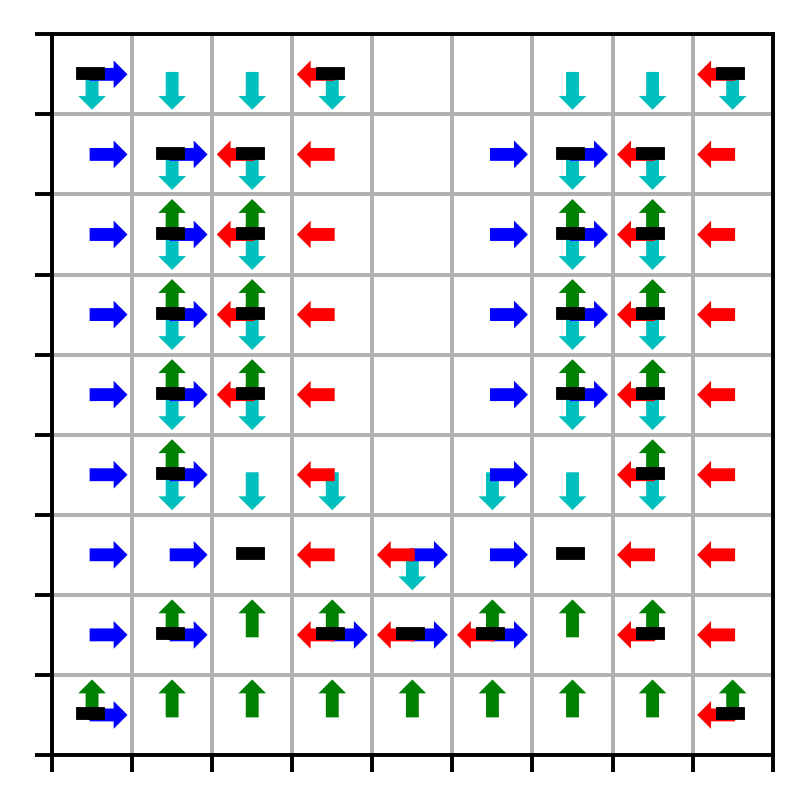}
        \label{fig:2c}
    }
    \subfigure[$\arg\min(Q^{min})$]
    {
        \includegraphics[width=0.46\hsize]{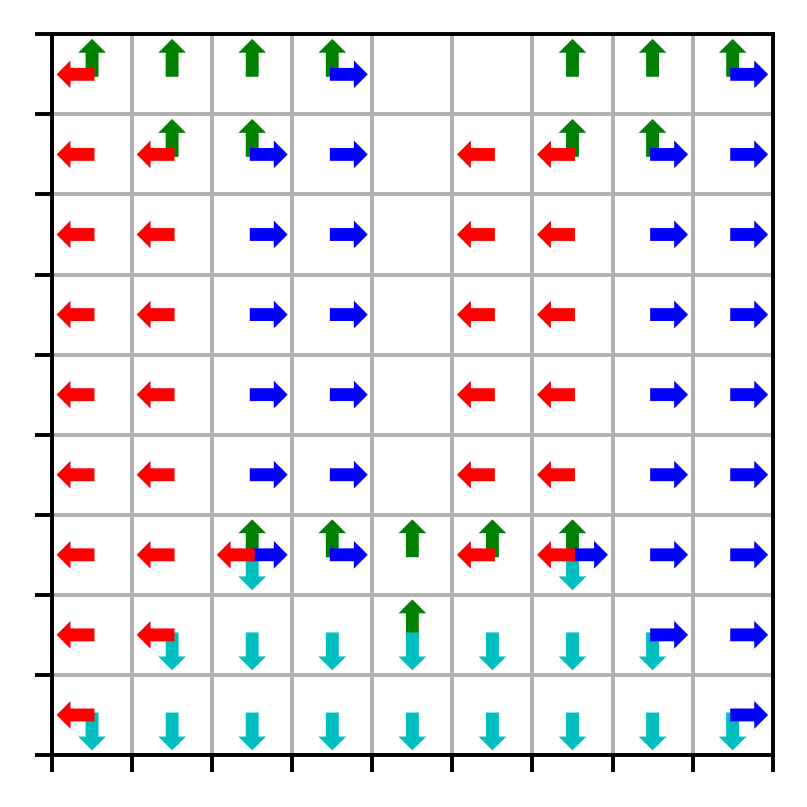}
        \label{fig:2d}
    }
    
    \caption{Two-end optimal policies derived from optimal action values $Q*$ by ``max'' and ``min'' operators with QVI in 9x9 U-maze Grid-world}
    \label{fig2}
    
\end{figure}

Fig.~\ref{fig1} shows the optimal state values obtained using ``max'' and ``min'' operators with QVI. The ``min'' operator (Fig.~\ref{fig:min}) learned more representative state values from the collision signals under the $\gamma=0.9$ setting. Three distinguishable areas emerged: the most painful states along the walls (a value of -1), the relatively safe zones (a value of -0.9), and the safest states (a value of -0.81). In contrast, the ``max'' operator (Fig.~\ref{fig:max}) only resulted in a uniform value of 0 everywhere. 

Fig.~\ref{fig2} suggests four types of optimal actions obtained from the learned values:

\begin{itemize}
\item[] (a) $\arg\max(Q^{max})$ - optimal policy
\item[] (b) $\arg\min(Q^{max})$ - flipped policy 
\item[] (c) $\arg\max(Q^{min})$ - flipped policy 
\item[] (d) $\arg\min(Q^{min})$ - optimal policy 
\end{itemize}

\noindent
where $Q^{max}$ represents the Q value obtained from updating with the ``max'' operator, and $Q^{min}$ corresponds to the ``min'' operator. Notably, both the optimal policy with the ``max'' operator (Fig.~\ref{fig:2a}) and the flipped policy with the ``min'' operator (Fig.~\ref{fig:2c}) exhibit pain-avoidance behaviors. Conversely, the flipped policy with a ``max'' operator (Fig.~\ref{fig:2b}) and the optimal policy with a ``min'' operator (Fig.~\ref{fig:2d}) showcase pain-seeking performances. The ``min'' operator tends to adopt long-sighted strategies; the ``max'' operator takes actions of only one step before encountering immediate pain. This observation is consistent with the insights from the different optimal state values they learned in Fig.~\ref{fig1}.

\begin{figure}
    \centering
    \subfigure[step length]
    {
        \includegraphics[width=0.472\hsize]{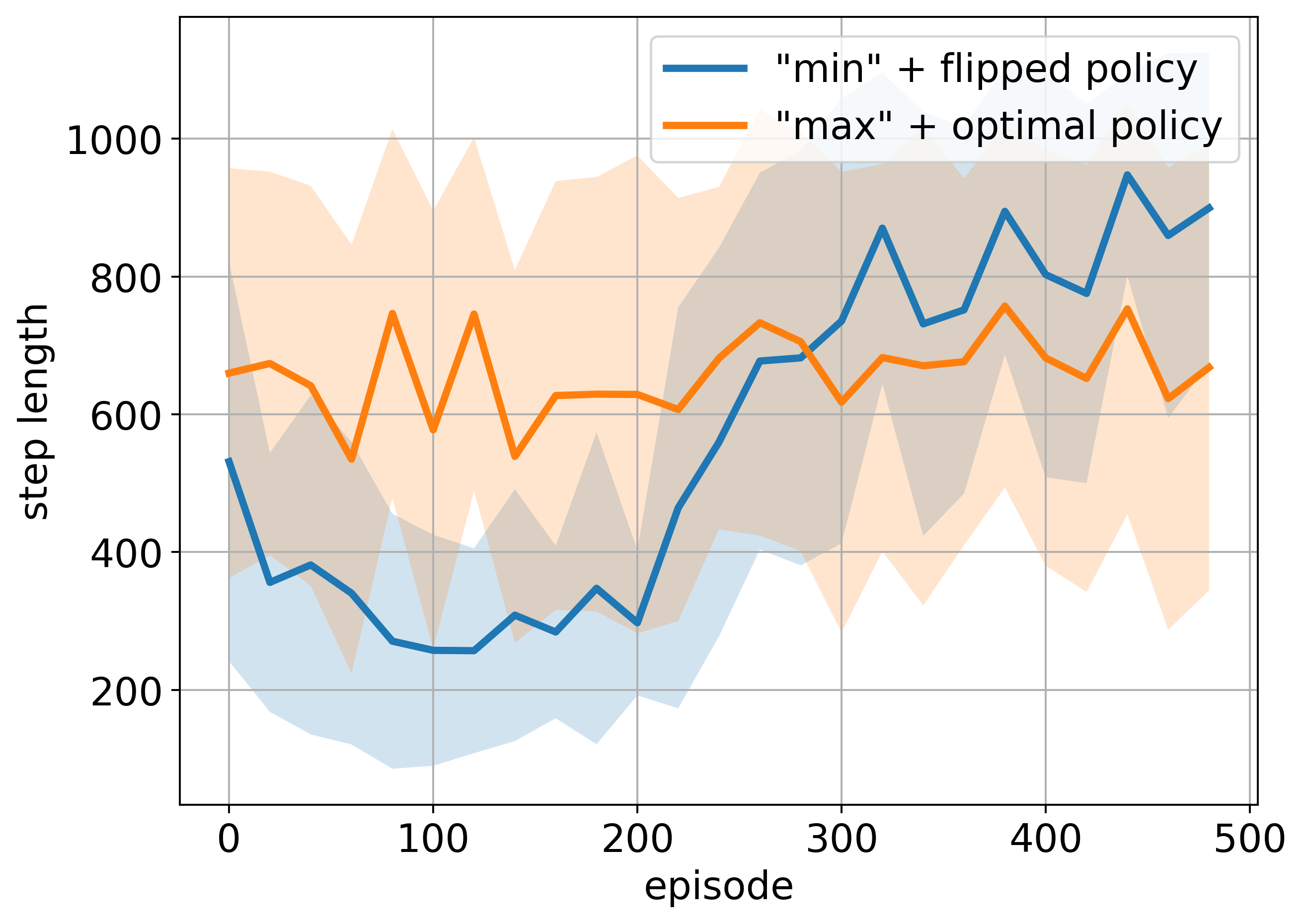}
        \label{fig:3a}
    }
    \subfigure[reward]
    {
        \includegraphics[width=0.46\hsize]{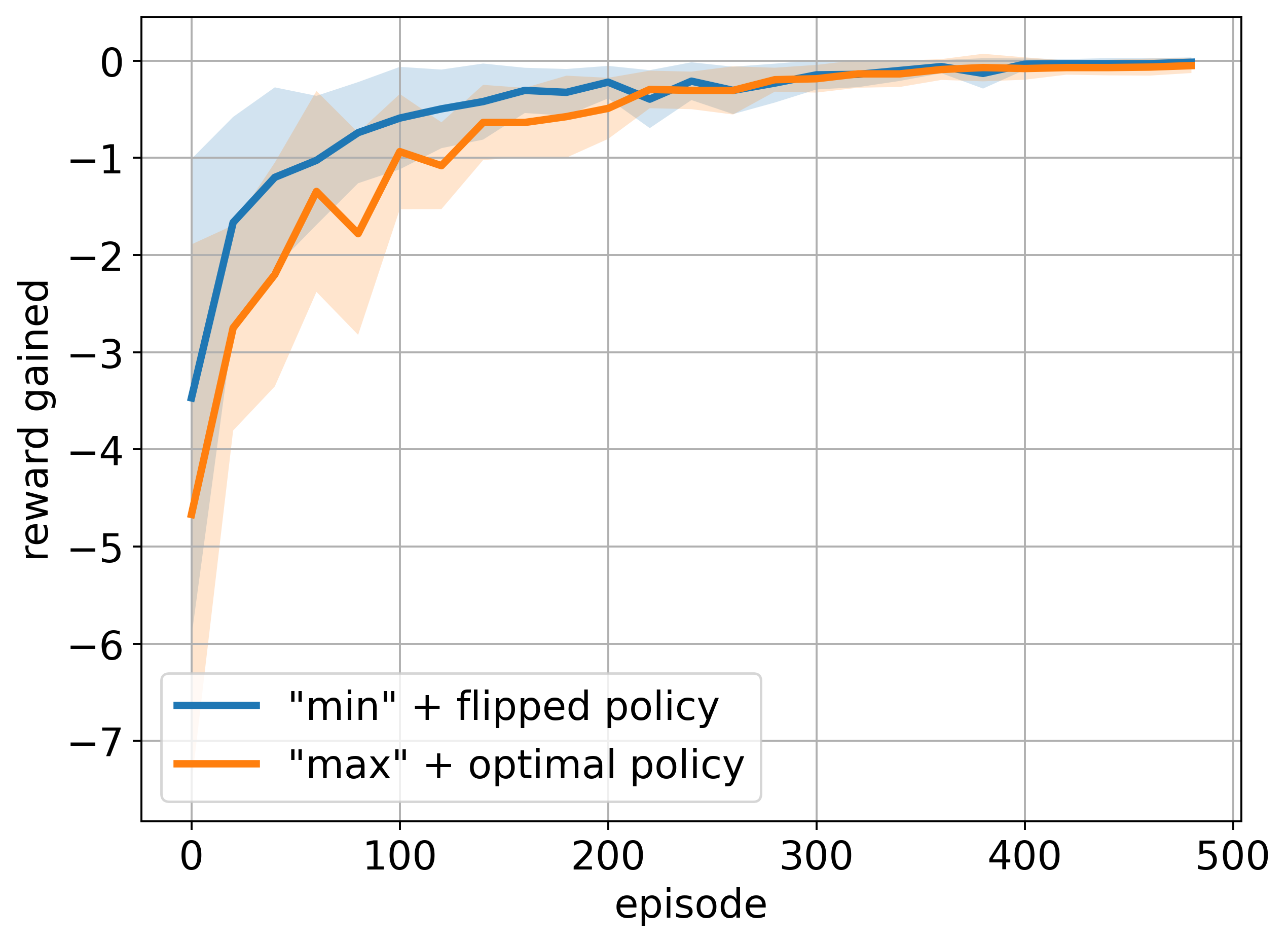}
        \label{fig:3b}
    }
    
    \caption{Smoothed Q-learning curves with ``min'' operator with flipped policy and ``max'' operator with optimal policy in terms of step lengths and rewards gained at each episode in 9x9 U-maze Grid-world}
    \label{fig3}
    
\end{figure}

We also conducted model-free Q-learning with the combinations of the ``max'' operator with the optimal policy and the ``min'' operator with the flipped policy in the same environment. Fig.~\ref{fig3} presents two smoothed learning curves, depicting the step length toward the goal and the reward gained in each episode against the episode length in a single run. Fig.~\ref{fig:3a} indicates that the ``max'' operator exhibited a random walk with high stochasticity throughout the entire learning, whereas the ``min'' operator learned to approach the goal quickly in the beginning and then identified two safe anchoring states to stay for a long time. This is consistent with the findings from QVI, as shown in Figs.~\ref{fig1} and ~\ref{fig2}. Fig.~\ref{fig:3b} suggests that the ``min'' operator with the flipped policy outperformed the ``max'' operator in reward gaining.

\begin{figure}[htbp]
\centerline{\includegraphics[width=1\hsize]{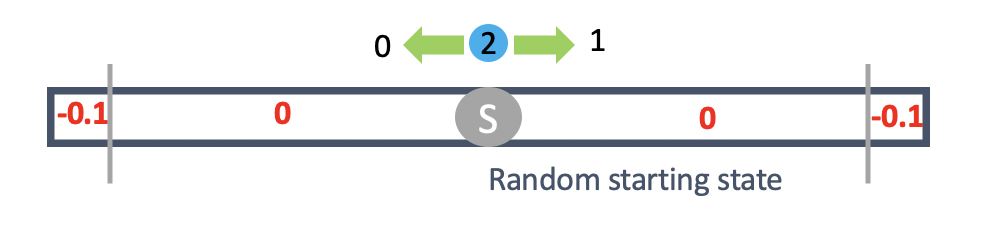}}
\caption{1x21 Chain Environment}
\label{chain}
\end{figure}

\begin{figure}[htbp]
\centerline{\includegraphics[width=1\hsize]{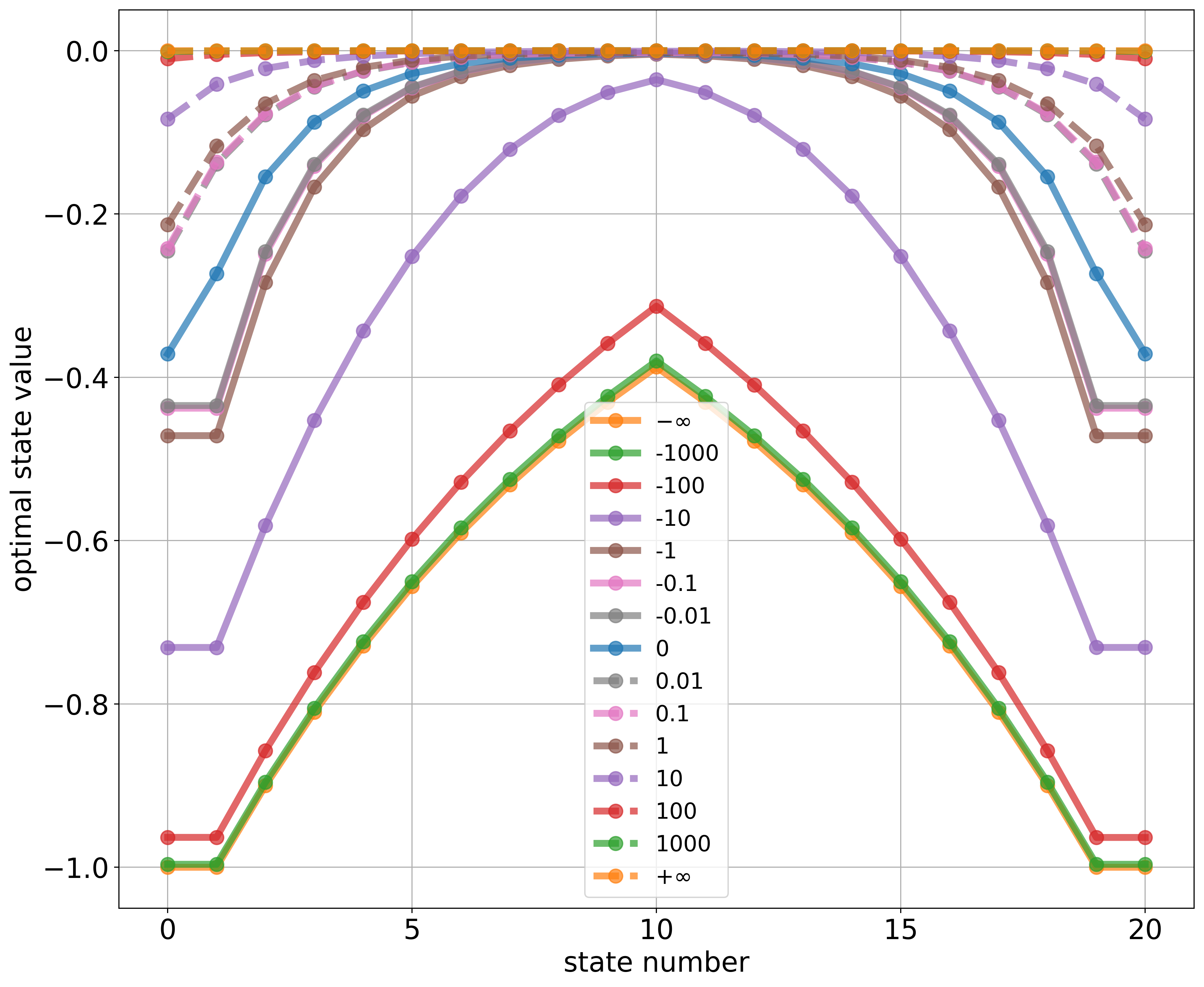}}
\caption{Optimal state values learned by SQL with entropy parameter $\eta \in \{-\infty, -1000,-100,-10,-1,-0.1,-0.01,0,0.01,0.1,$ $1,10,100,1000,\infty\}$ in 1x21 Chain environment}
\label{fig5}
\end{figure}

Next we present a range of operators governed by $\eta \in$ $\{-\infty, -1000, -100, -10, -1, -0.1, -0.01, 0, 0.01, 0.1, 1, 10,$ $100,1000,\infty\}$ 
with SQL applied in a 1x21 Chain environment (Fig.~\ref{chain}). We set a negative reward of $r=-0.1$ on both edge states of the chain, and let the agent start to move randomly. The agent has two directional actions and one ``stop'' action. There are no absorbing states, which means the agent stops learning when it reaches the episode length. Note that as shown in Eq.(\ref{eq:mmq}), $\eta \geq 0$ corresponds to ``max'', ``mellow-max'' or ``mean'' operators with non-flipped policies, and $\eta<0$ corresponds to ``min'' or ``mellow-min'' operators with flipped policies.

Fig.~\ref{fig5} displays the optimal state values against 21 states. The solid lines represent the negative and zero $\eta$s with flipped behavior policies, while the dashed lines are positive $\eta$s with optimal policies. This result suggests that optimal state values with the negative $\eta$s (combinations of ``min'' or ``mellow-min'' operators and ``flipped'' policies) have larger variances than those with the non-negative $\eta$s (combinations of ``max'', ``mellow-max'', or ``mean'' operators with optimal policies). The former covers a broader range of values back-propagated from the negative signals than the latter. As $\eta$ approaches $-\infty$, the gradient of the optimal value becomes steeper with respect to the states from the edge to the middle. For positive $\eta$s, smaller $\eta$s result in a more pronounced bend near the edge states compared to the larger $\eta$s. This suggests the smoothness of the ``max'' operator is able to convey dangerous information; the smoothness of the ``min'' operator can fine-tune the magnitude of the negative signal to be learned.  

\subsection{Gazebo Navigation}

\begin{figure}[htbp]
\centerline{\includegraphics[width=1\hsize]{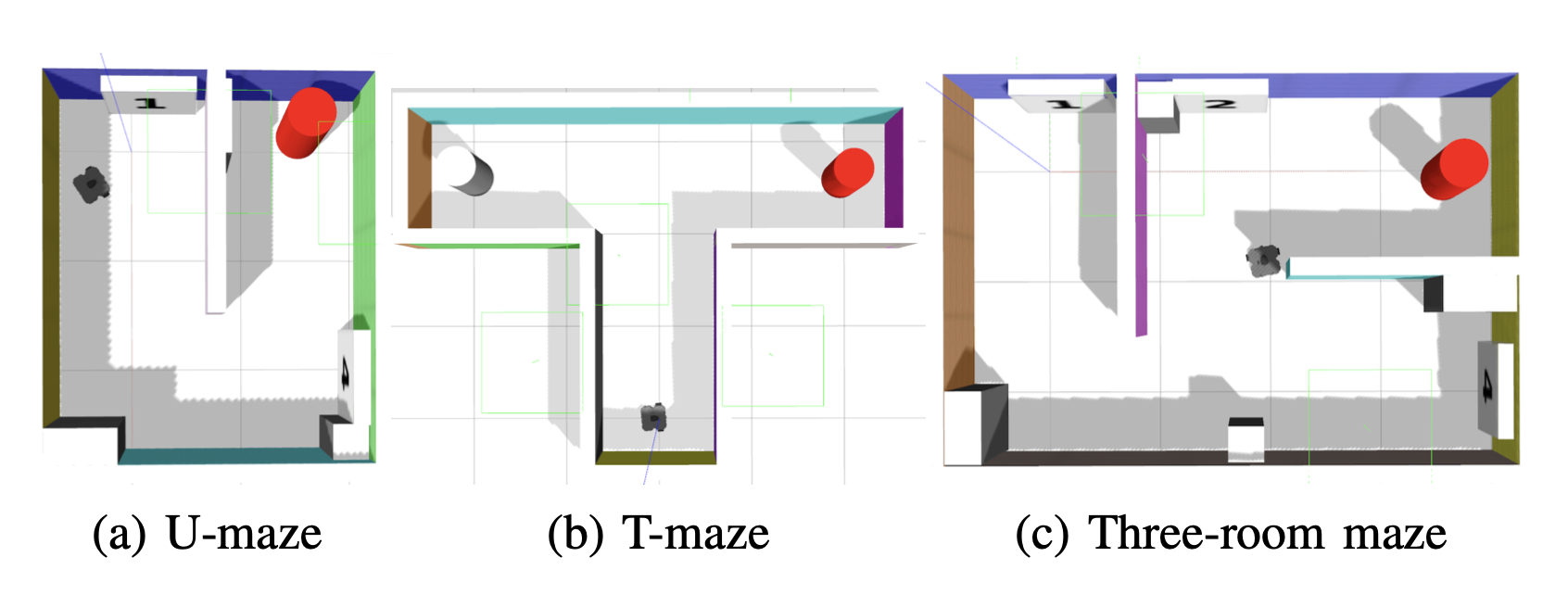}}
\caption{Three different mazes}
\label{fig6}
\end{figure}

We validated the softDMPs in three types of maze-solving environments using Turtlebot3 Waffle Pi under the ROS Gazebo simulator (Fig.~\ref{fig6}). The environments were previously evaluated in \cite{Wang2021a}, and the experimental setting were identical to the descriptions in Section 4.2.1 \cite{Wang2021a}. Specifically, the environment features five directional actions, where the agent receives a reward of $r=+5$ at the absorbing-goal state (the red cylinder) and a reward of $r=-0.5$ upon colliding with a wall. We adopted the same fusion network with with RGB Image and Lidar inputs as shown in Fig.9 of \cite{Wang2021a}.

\begin{figure}
    \centering
    \subfigure[U-maze scattered metrics]
    {
        \includegraphics[width=0.46\hsize]{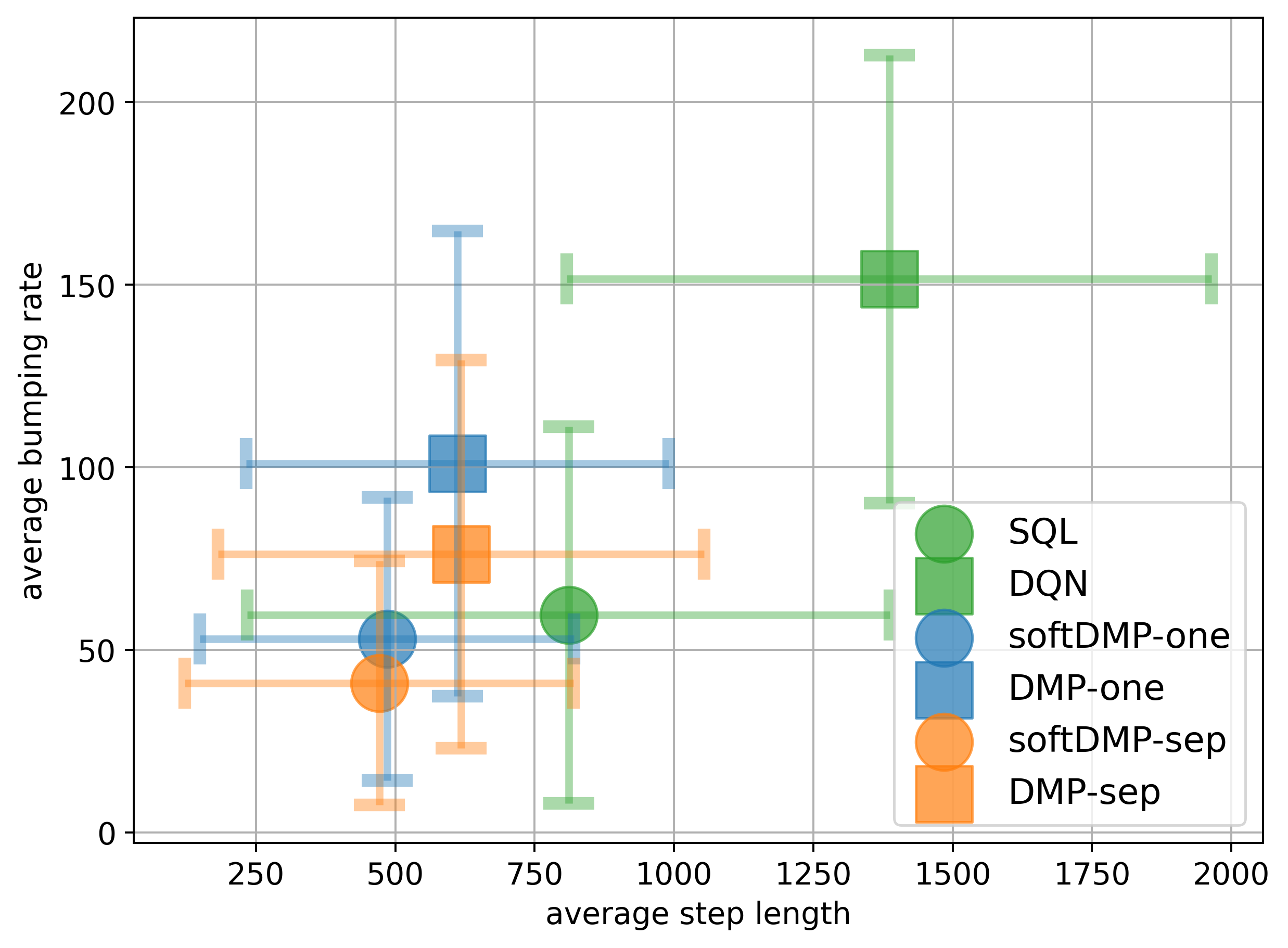}
        \label{fig:uscatter}
    }
    \subfigure[U-maze learning curves]
    {
        \includegraphics[width=0.47\hsize]{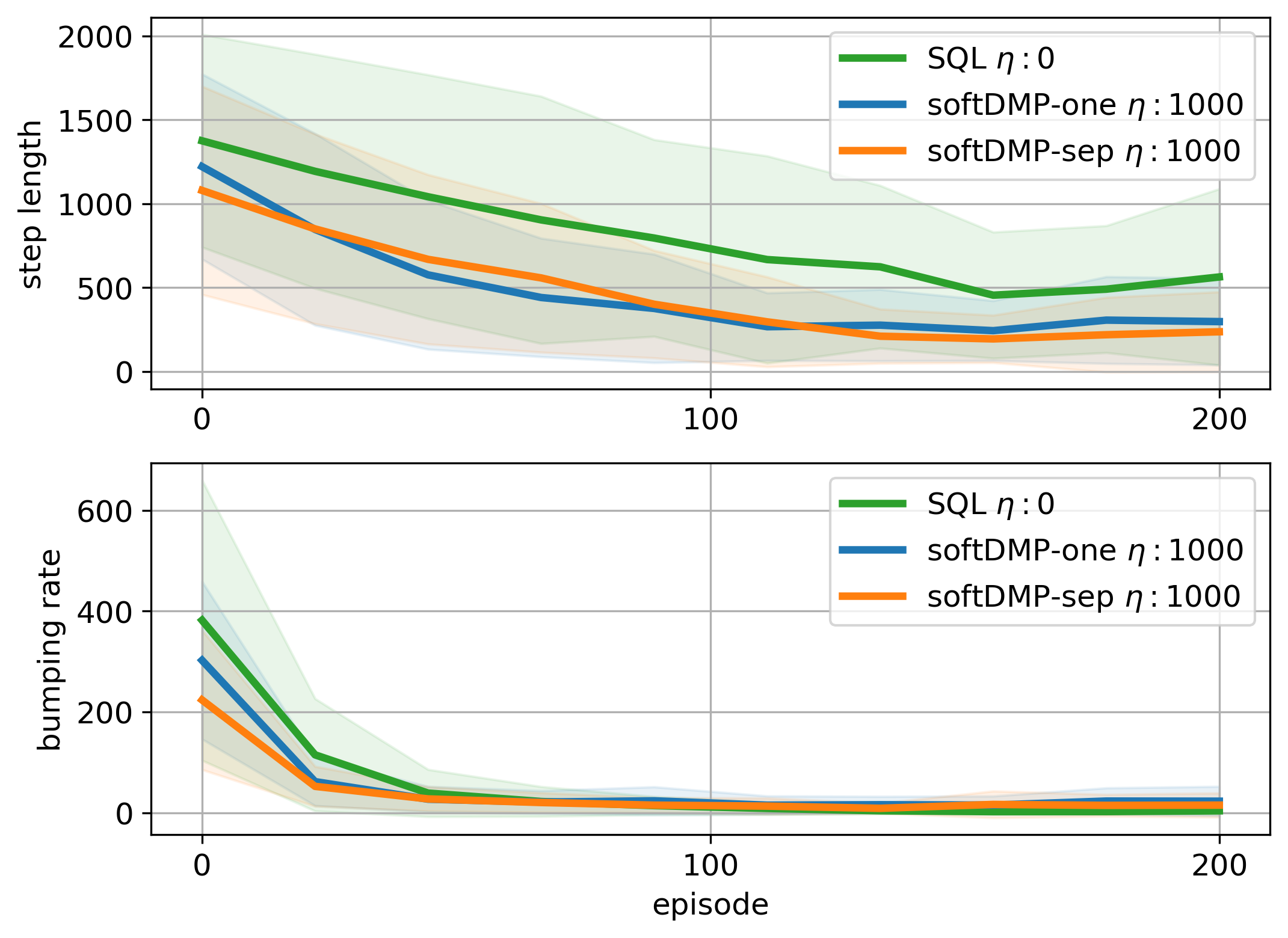}
        \label{fig:ucurve}
    }
    \subfigure[T-maze scattered metrics]
    {
        \includegraphics[width=0.46\hsize]{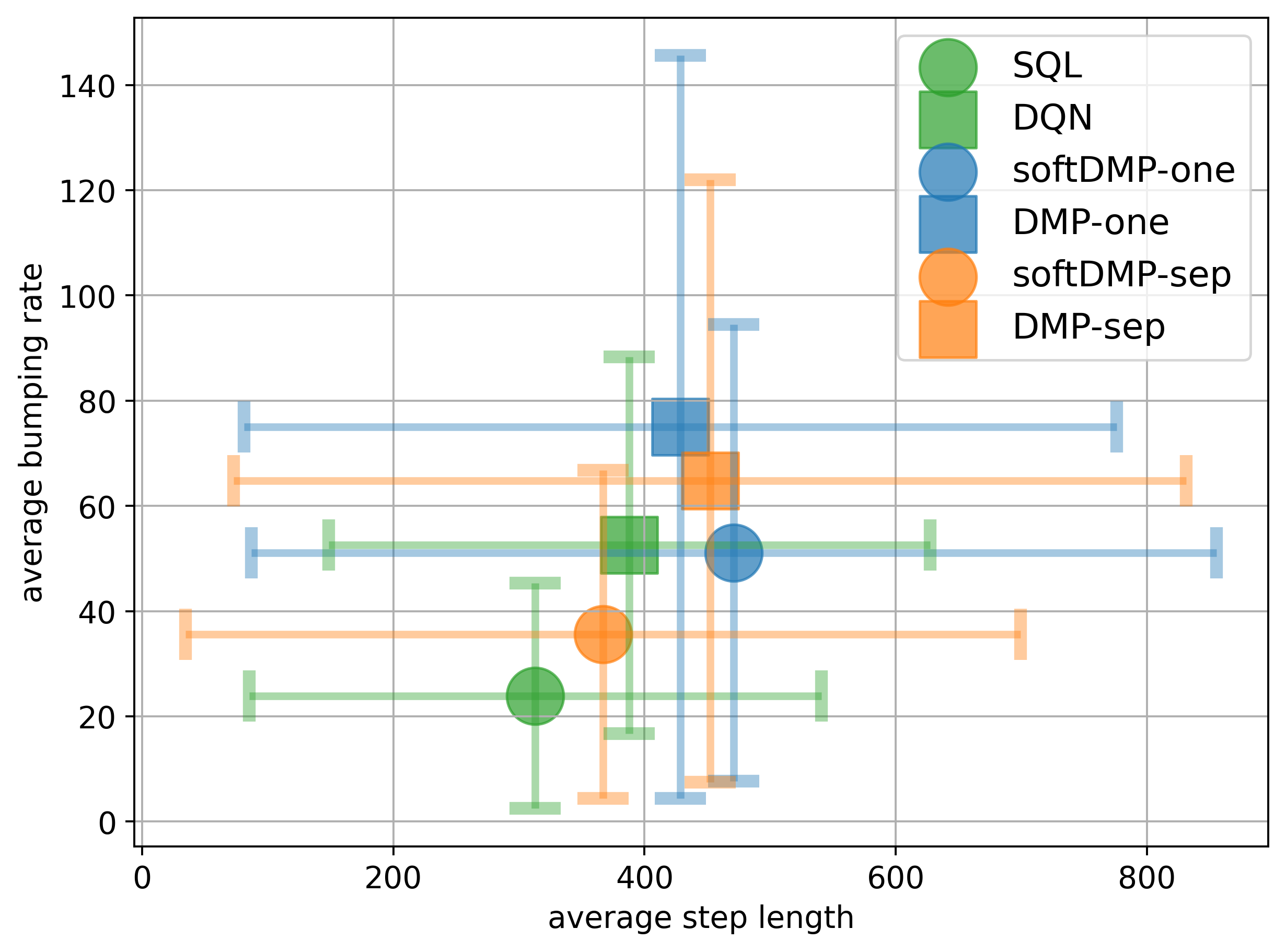}
        \label{fig:tscatter}
    }
    \subfigure[T-maze learning curves]
    {
        \includegraphics[width=0.47\hsize]{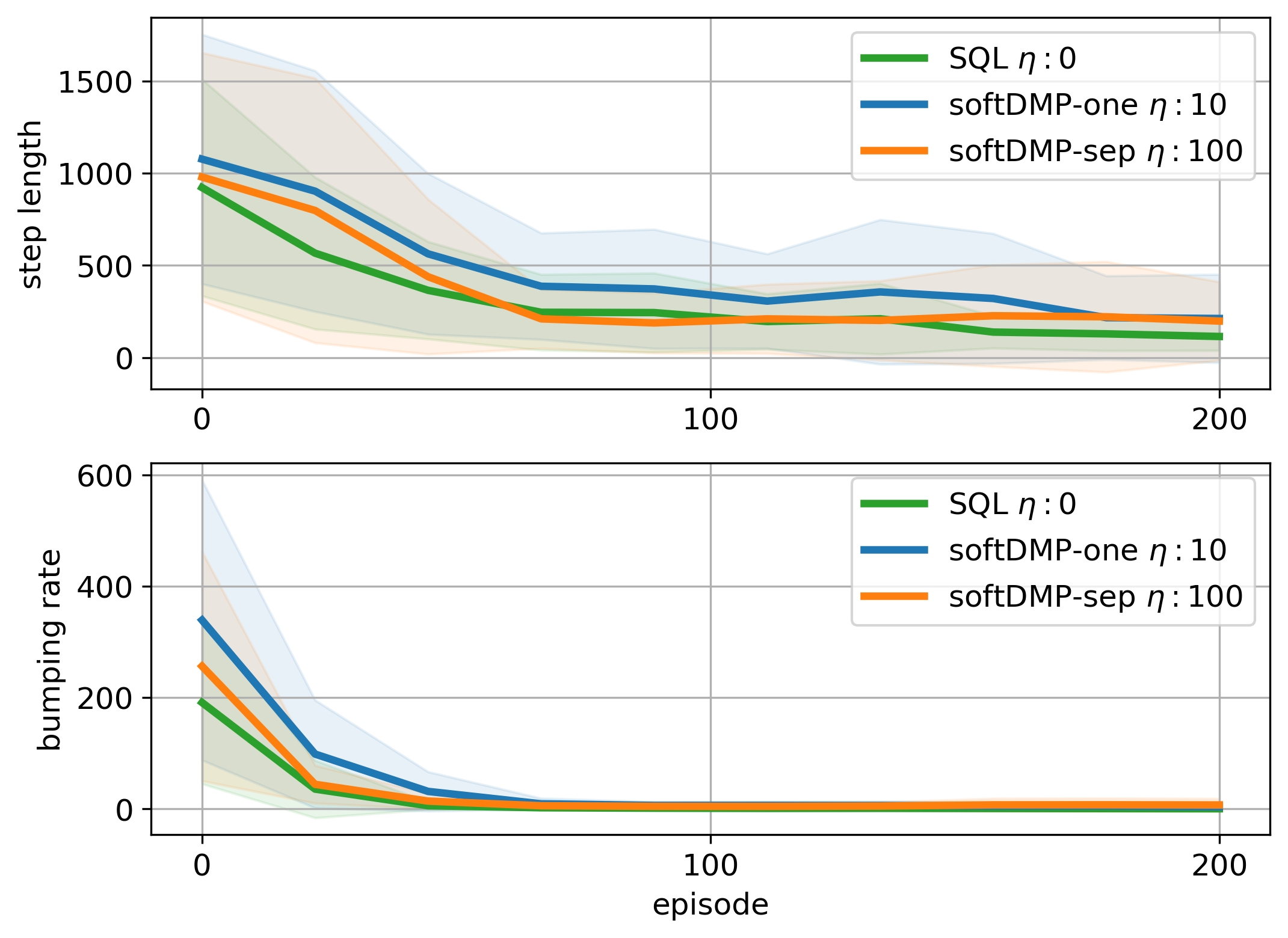}
        \label{fig:tcurve}
    }
    \subfigure[Three-room scattered metrics]
    {
        \includegraphics[width=0.46\hsize]{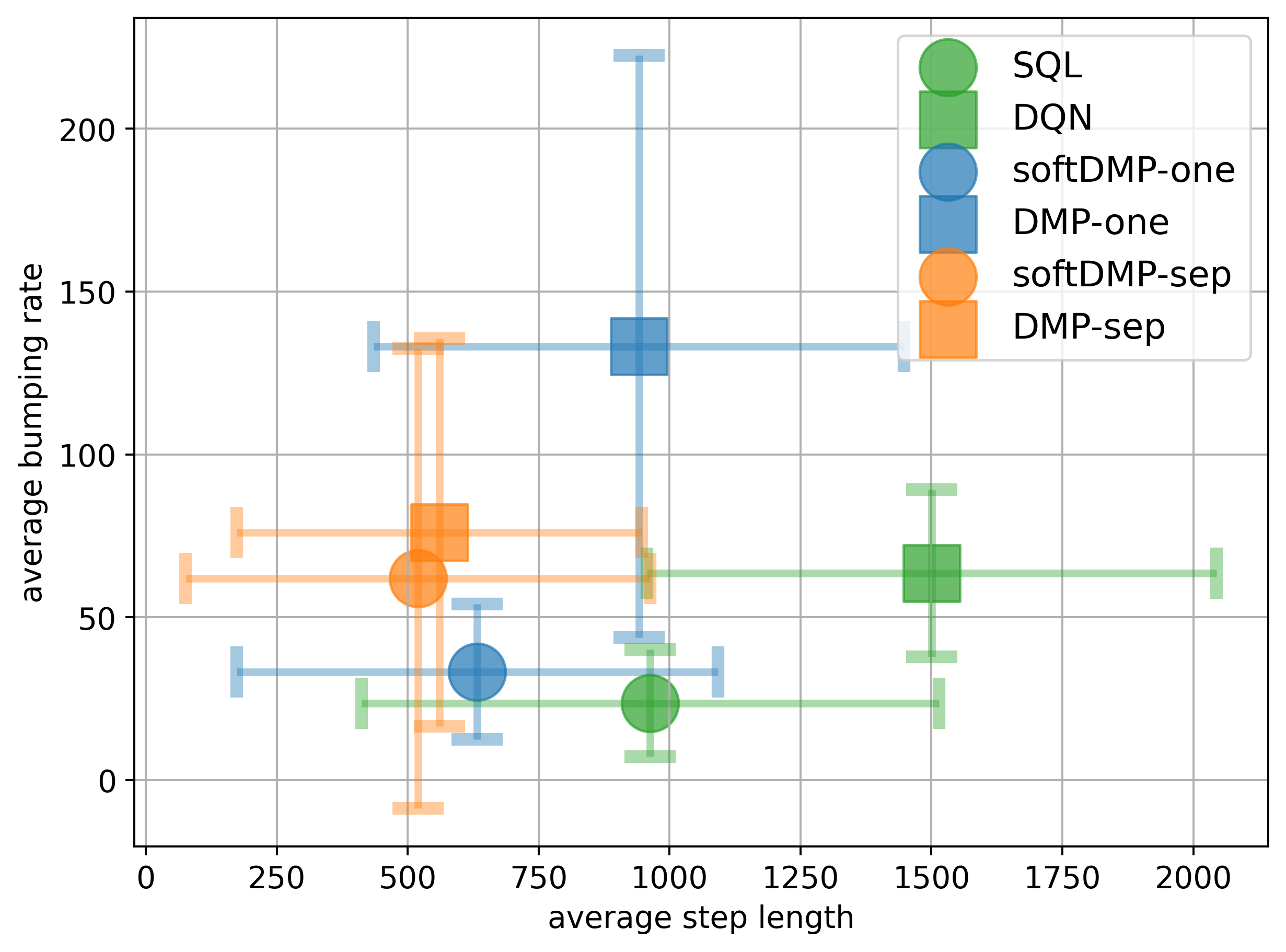}
        \label{fig:3rscatter}
    }
    \subfigure[Three-room learning curves]
    {
        \includegraphics[width=0.47\hsize]{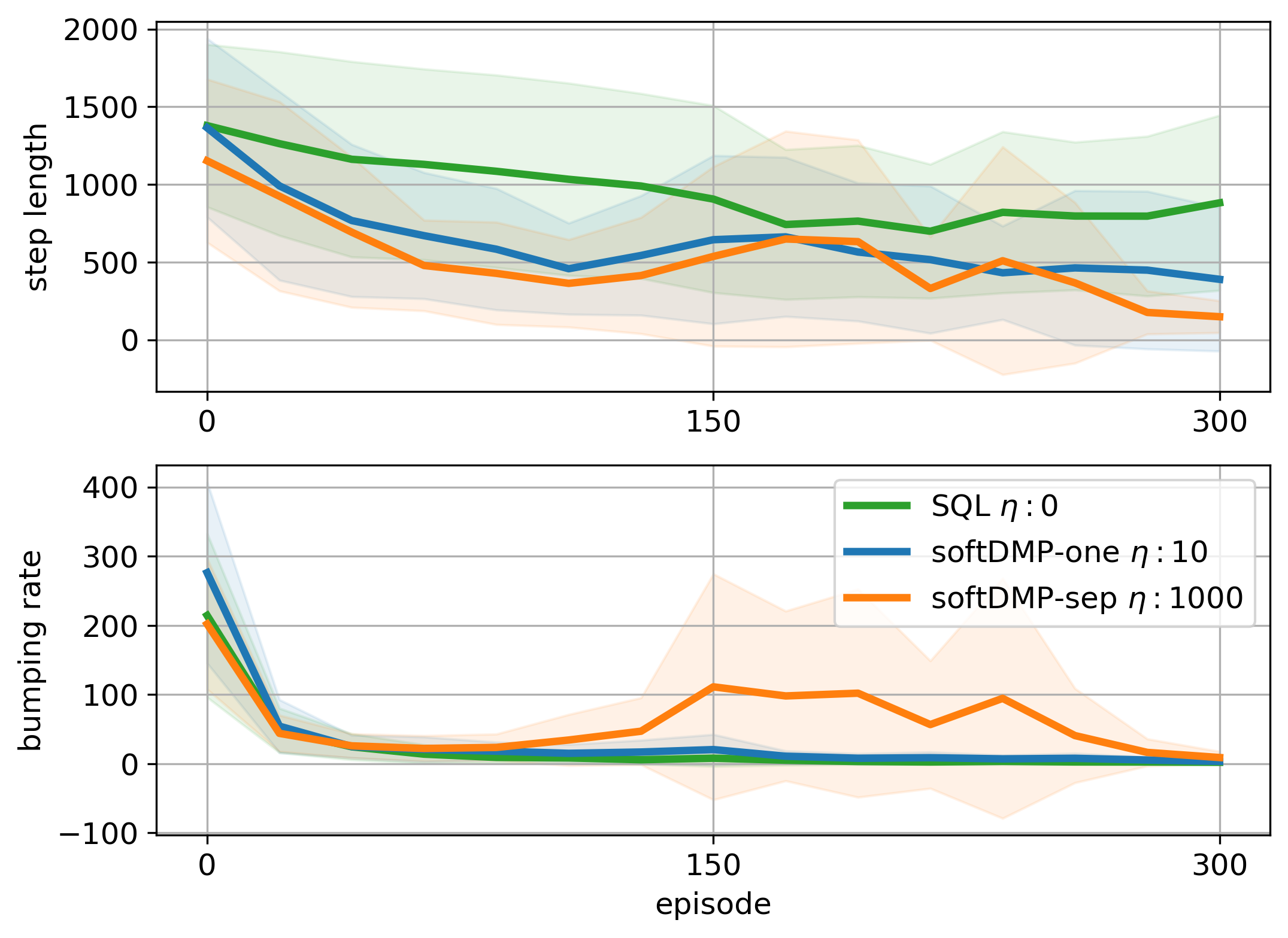}
        \label{fig:3rcurve}
    }
    \caption{Learning performances of each method in terms of step lengths and bumping rates of three mazes}
    \label{fig7}
    
\end{figure}

Our baselines include DQN ($\eta=\infty$ in the SQL case), SQL with $\eta \in \{10000,1000,100,10,0\}$, and DMP ($(\eta_+,\eta_-)=(+\infty,-\infty)$ in the softDMP case). All the reward-punishment related methods adopted a hard-max weighting scheme for mixing two sub-policies as a behavior policy \cite{Wang2021a}. Note that hard-max weighting is based on real-time comparisons of $V_+(s)$ and $|V_-(s)|$. Since the new objectives comprise the entropy term, the constraint of $\gamma_+$ and $\gamma_-$ with the reward setting is slightly violated, as discussed in Section 3.5 \cite{Wang2021a}. We tested softDMP with a separate buffer (softDMP-sep) and with only one buffer (softDMP-one). The entropy parameter pair $(\eta_+,\eta_-)$ for each maze was set as follows: $(10000,-10000)$, $(1000,-1000)$, $(100,-100)$, $(10,-10)$, and $(0,0)$. The RL hyperparameters were $\alpha=0.025$ and $\gamma=0.9$ for the DQN and SQLs, $(\alpha_+,\alpha_-)=(0.025,0.001)$ and $(\gamma_+,\gamma_-)=(0.99,0.9)$ for the DMPs and the softDMPs in terms of the U-maze and T-maze. In the case of the three-room maze, $\alpha=0.03$ and $\alpha_+=0.03$, while the others remained the same. We observed that the relative performances of each method were hyper-parameter sensitive.

Fig.~\ref{fig7} shows the learning performances of each method in the three mazes. Figs.~\ref{fig:uscatter},~\ref{fig:tscatter}, and ~\ref{fig:3rscatter} present scattered metrics in terms of average step lengths and collisions, where being further to the lower left corner indicates better overall performance. Figs.~\ref{fig:ucurve},~\ref{fig:tcurve}, and ~\ref{fig:3rcurve} demonstrate the averaged learning curves of the step lengths and collisions against the episode length. The results were averaged from five trials for each method in each maze. Note that the best performance of $\eta$ was selected to display for SQLs and softDMPs. They are: $0$ for SQL, $1000$ for softDMP-sep and softDMP-one for U-maze; $0$ for SQL, $100$ for softDMP-sep and $10$ for softDMP-one for T-maze; $0$ for SQL, $1000$ for softDMP-sep and $10$ for softDMP-one for the three-room maze. From Figs.~\ref{fig:uscatter},~\ref{fig:tscatter}, and ~\ref{fig:3rscatter}, we conclude the following general tendencies: SQLs and softDMPs outperform their DQN and DMP prototypes in both two metrics. Additionally, softDMP-seps outperform softDMP-ones in most cases, except in the three-room maze, where softDMP-sep achieves shorter step lengths towards the goal, while softDMP-one achieves fewer collision rates. SQL with $\eta=0$ acquired impressive collision avoidance capabilities in the T-maze and the three-room maze. This result is attributed to the fact that SQL with the ``mean'' operator does not exhibit the extreme behaviors seen with ``max'' and ``min''. The ``max'' operator tends to prioritize the shortest path towards the goal, often disregarding potential dangers, whereas the ``min'' operator errs on the side of caution, often excessively favoring the safest states. In contrast, the ``mean'' operator strikes a balance, seeking out a relatively safer path while mitigating potential risks to some extent. In terms of step length in the case of the three-room maze, softDMP-sep experienced a steeper decline in the early stage, a hump in the middle, and a convergence in the end stage of learning than the other methods. This suggests that a separate buffer might assist quick learning of $Q^-$ in the beginning. When the agent's attention turns to goal-achieving, less and less effective experience can be collected for updating $Q^-$. The approximation of $Q^-$ might face overfitting in the middle of learning and finally reach a convergence due to the domination of the positive sub-module being selected by the hard-max weighting scheme. 

\section{Conclusions}

We introduced a maximum entropy technique into a reward-punishment RL framework for the development of soft Deep MaxPain algorithms (softDMPs). We delved into why ``min'' and ``mellow-min'' operators, in collaboration with ``flipped'' policies, enhance learning performance, especially in dealing with negative rewards. We applied these insights along with the soft treatment on optimization operators provided by maximum entropy to the realm of deep reinforcement learning and proposed a separate buffer technique to address the experience shortage for learning negative action values. Our methods outperformed the previous methods: DQN, SQL and DMPs and achieved a competitive performance in terms of quickly locating the shortest path to the goal and persuasive collision avoidance in real-time vision-based navigation within the Gazebo maze environments using Turtlebot3. In the future, we will explore further properties of entropy parameter $\eta$, investigate the overfitting problem of the negative action-value $Q^-$, and relate it to offline RL.

\section*{Acknowledgments}
This work was supported in part by the New Energy and Industrial Technology Development Organization through the results obtained from Project JPNP20006 and in part by JSPS KAKENHI under Grant JP21H03527.

\bibliographystyle{unsrt}
\bibliography{myreference}

\begin{thebibliography}{10}

\bibitem{Elfwing2017a}
S.~Elfwing and B.~Seymour.
\newblock Parallel reward and punishment control in humans and robots: safe
  reinforcement learning using the maxpain algorithm.
\newblock In {\em Proc. of the 7th Joint IEEE International Conference on
  Development and Learning and on Epigenetic Robotics}, pages 140--7, 2017.

\bibitem{Seymour2007a}
B.~Seymour, N.~Daw, P.~Dayan, T.~Singer, and R.~Dolan.
\newblock Differential encoding of losses and gains in the human striatum.
\newblock {\em Journal of Neuroscience}, 27(18):4826--31, 2007.

\bibitem{Seymour2012a}
B.~Seymour, N.~Daw, J.~P. Roiser, P.~Dayan, and R.~Dolan.
\newblock Serotonin selectively modulates reward value in human
  decision-making.
\newblock {\em Journal of Neuroscience}, 32(17):5833--42, 2012.

\bibitem{Eldar2016a}
E.~Eldar, T.~U. Hauser, P.~Dayan, and R.~J. Dolan.
\newblock Striatal structure and function predict individual biases in learning
  to avoid pain.
\newblock {\em Proceedings of the National Academy of Sciences of the United
  States of America}, 113(17):4812--7, 2016.

\bibitem{Wang2017b}
J.~Wang, S.~Elfwing, and E.~Uchibe.
\newblock Deep reinforcement learning by parallelizing reward and punishment
  using the maxpain architecture.
\newblock In {\em Proc. of the 8th Joint IEEE International Conference on
  Development and Learning and on Epigenetic Robotics}. IEEE, 2018.

\bibitem{Wang2021a}
J.~Wang, S.~Elfwing, and E.~Uchibe.
\newblock Modular deep reinforcement learning from reward and punishment for
  robot navigation.
\newblock {\em Neural Networks}, 135:115--26, 2021.

\bibitem{Asadi2017a}
K.~Asadi and M.~L. Littman.
\newblock An alternative softmax operator for reinforcement learning.
\newblock In {\em Proc. of the 34th International Conference on Machine
  Learning}, 2017.

\bibitem{Goodfellow2014a}
I.~Goodfellow, J.~Pouget-Abadie, M.~Mirza, B.~Xu, D.~Warde-Farley, S.~Ozair,
  A.~Courville, and Y.~Bengio.
\newblock Generative adversarial nets.
\newblock In {\em Advances in Neural Information Processing Systems 27}, pages
  2672--80, 2014.

\bibitem{cql}
S.~P. Singh.
\newblock Transfer of learning by composing solutions of elemental sequential
  tasks.
\newblock {\em Machine Learning}, 8(3-4):323--339, 1992.

\bibitem{gmq}
J.~Karlsson.
\newblock {\em Learning to solve multiple goals}.
\newblock PhD thesis, University of Rochester, 1997.

\bibitem{hra}
H.~van Seijen, M.~Fatemi, J.~Romoff, R.~Laroche, T.~Barnes, and J.~Tsang.
\newblock Hybrid reward architecture for reinforcement learning.
\newblock In {\em Advances in Neural Information Processing Systems 30}, 2017.

\bibitem{humphrys1996}
M.~Humphrys.
\newblock Action selection methods using reinforcement learning.
\newblock In {\em From Animals to Animats 4: Proceedings of the Fourth
  International Conference on Simulation of Adaptive Behavior}, pages 135--144,
  1996.

\bibitem{doya2002}
K.~Doya, K.~Samejima, K.~Katagiri, and M.~Kawato.
\newblock Multiple model-based reinforcement learning.
\newblock {\em Neural Computation}, 14(6):1347--1369, 2002.

\bibitem{okada2001two}
H.~Okada, H.~Yamakawa, and T.~Omori.
\newblock Two dimensional evaluation reinforcement learning.
\newblock In {\em International Work-Conference on Artificial Neural Networks},
  pages 370--377. Springer, 2001.

\bibitem{lowe2013exploring}
R.~Lowe and T.~Ziemke.
\newblock Exploring the relationship of reward and punishment in reinforcement
  learning.
\newblock In {\em Proc. of the 2013 IEEE Symposium on Adaptive Dynamic
  Programming and Reinforcement Learning (ADPRL)}, pages 140--147. IEEE, 2013.

\bibitem{rlac}
T.~Kobayashi, T.~Aotani, J.~R. Guadarrama-Olvera, E.~Dean-Leon, and G.~Cheng.
\newblock Reward-punishment actor-critic algorithm applying to robotic
  non-grasping manipulation.
\newblock In {\em 2019 Joint IEEE 9th International Conference on Development
  and Learning and Epigenetic Robotics (ICDL-EpiRob)}, pages 37--42. IEEE,
  2019.

\bibitem{split}
B.~Lin, G.~A. Cecchi, D.~Bouneffouf, J.~Reinen, and I.~Rish.
\newblock A story of two streams: Reinforcement learning models from human
  behavior and neuropsychiatry.
\newblock In {\em Proc. of the 19th International Conference on Autonomous
  Agents and Multi-Agent Systems}, pages 744--752, 2020.

\bibitem{Fox2016a}
R.~Fox, A.~Pakman, and N.~Tishby.
\newblock Taming the noise in reinforcement learning via soft updates.
\newblock In {\em Proc. of the 32nd Conference on Uncertainty in Artificial
  Intelligence}, 2016.

\bibitem{Azar2012a}
M.~G. Azar, V.~G\'omez, and H.~J. Kappen.
\newblock Dynamic policy programming.
\newblock {\em Journal of Machine Learning Research}, 13:3207--45, 2012.

\bibitem{Toussaint2009a}
M.~Toussaint.
\newblock Robot trajectory optimization using approximate inference.
\newblock In {\em Proc. of the 26th International Conference on Machine
  Learning}, pages 1049--56, 2009.

\bibitem{Haarnoja2017a}
T.~Haarnoja, H.~Tang, P.~Abbeel, and S.~Levine.
\newblock Reinforcement learning with deep energy-based policies.
\newblock In {\em Proc. of the 34th International Conference on Machine
  Learning}, pages 1352--61, 2017.

\bibitem{Haarnoja2018a}
T.~Haarnoja, A.~Zhou, P.~Abbeel, and S.~Levine.
\newblock Soft actor-critic: Off-policy maximum entropy deep reinforcement
  learning with a stochastic actor.
\newblock In {\em Proc. of the 35th International Conference on Machine
  Learning}, pages 1861--70, 2018.

\bibitem{Levine2018a}
L.~Sergey.
\newblock Reinforcement learning and control as probabilistic inference:
  Tutorial and review.
\newblock {\em CoRR, abs/1805.00909}, 2018.

\bibitem{Kozuno2019a}
T.~Kozuno, E.~Uchibe, and K.~Doya.
\newblock Theoretical analysis of efficiency and robustness of softmax and
  gap-increasing operators in reinforcement learning.
\newblock In {\em Proc. of the 22nd International Conference on Artificial
  Intelligence and Statistics}, pages 2695--3003, 2019.

\bibitem{Eysenbach2022a}
B.~Eysenbach and S.~Levine.
\newblock Maximum entropy rl (provably) solves some robust rl problems.
\newblock In {\em Proc. of the 10th International Conference on Learning and
  Representations}, 2022.

\end{thebibliography}

\end{document}